\pdfoutput = 1
\documentclass[twocolumn]{article}
\usepackage[numbers]{natbib}

\usepackage{style}

\usepackage{gensymb}
\usepackage{mhchem}

\usepackage{xurl}

\usepackage{times}

\usepackage[english]{babel}
\usepackage{blindtext}
\usepackage{graphicx}
\usepackage{amsmath, amsthm, amssymb, bbm, bm}
\usepackage{enumerate}
\usepackage{float}      % figure inside minipage,
\usepackage{subcaption}  % does not go well with subfig.
\usepackage{wrapfig}
\usepackage[margin=0cm]{caption}
\usepackage{tabularx}

\usepackage{multirow}
\usepackage{hhline}
\usepackage{makecell}
\usepackage{placeins}  % stop floating

\usepackage[x11names, usenames, dvipsnames, svgnames, table]{xcolor}
\definecolor{firebrick}{rgb}{.698,.133,.133}
\definecolor{mybluelight}{rgb}{0.9, 0.9, 1.}
\definecolor{myorangelight}{rgb}{1., 0.9, 0.9}

\usepackage[utf8]{inputenc} % allow utf-8 input
\usepackage[T1]{fontenc}    % use 8-bit T1 fonts
\usepackage{url}            % simple URL typesetting
\usepackage{booktabs, colortbl}       % professional-quality tables
\usepackage{amsfonts}       % blackboard math symbols
\usepackage{nicefrac}       % compact symbols for 1/2, etc.
\usepackage{microtype}      % microtypography

\usepackage{csquotes}
\usepackage{latexsym}

\usepackage{pifont}% http://ctan.org/pkg/pifont
\usepackage[boxruled, vlined, linesnumbered]{algorithm2e}
\SetAlFnt{\small}
\SetAlCapFnt{\small}
\SetAlCapNameFnt{\small}
\usepackage{algorithmic}
\algsetup{linenosize=\tiny}

\let\oldnl\nl% Store \nl in \oldnl
\newcommand{\nonl}{\renewcommand{\nl}{\let\nl\oldnl}}% Remove line number for one line

\usepackage{paralist}

\usepackage{xspace}
\usepackage{soul}
\usepackage{dsfont}
\usepackage{stmaryrd}
\usepackage[textwidth=15mm]{todonotes}
\usepackage{dirtytalk}
\usepackage{pbox}
\usepackage{cprotect}

\usepackage{verbatim}
\usepackage{textcomp}
\usepackage[normalem]{ulem}

\usepackage{mathtools}
\usepackage{etextools}
\usepackage[inline]{enumitem}

\usepackage[colorlinks=true,allcolors=firebrick,bookmarks=false]{hyperref}

\usepackage[toc,page,header]{appendix}
\usepackage{minitoc}

\renewcommand \thepart{}

\newcommand\war{\texttt{WAR}\xspace}

\definecolor{darkergreen}{RGB}{21, 152, 56}
\definecolor{red2}{RGB}{252, 54, 65}
\definecolor{Gray}{gray}{0.85}
\newcolumntype{g}{>{\columncolor{Gray}}c}

\let\OLDthebibliography\thebibliography
\renewcommand\thebibliography[1]{
  \OLDthebibliography{#1}
  \setlength{\parskip}{0pt}
  \setlength{\itemsep}{0pt plus 0.3ex}
}
\theoremstyle{definition}

\DeclarePairedDelimiterX{\divx}[2]{(}{)}{%
  #1\;\delimsize\|\;#2%
}

\makeatletter
\@namedef{ver@everyshi.sty}{}
\newcommand{\removelatexerror}{\let\@latex@error\@gobble}

\newcommand\bah{\texttt{BAH}\xspace}
\newcommand\fonescore{\texttt{F1}\xspace}
\newcommand\wfonescore{\texttt{WF1}\xspace}
\newcommand\avgfonescore{\texttt{AVGF1}\xspace}
\newcommand\apscore{\texttt{AP}\xspace}

\newcommand\rafdb{\texttt{RAF-DB}\xspace}
\newcommand\affectnet{\texttt{AffectNet}\xspace}
\newcommand\affwildtwo{\texttt{Aff-wild2}\xspace}

\newlength{\NFwidth}
\NewDocumentCommand{\NFline}{O{l}m}{\footnotesize\makebox[\NFwidth][#1]{#2}}

\NewDocumentCommand{\NFentry}{sm}{%
  \makebox[.5\NFwidth][l]{\normalsize
    \IfBooleanT{#1}{\makebox[0pt][r]{\textbullet\ }}%
    #2}\ignorespaces}

\newcommand\blfootnote[1]{%
  \begingroup
  \renewcommand\thefootnote{}\footnote{#1}%
  \addtocounter{footnote}{-1}%
  \endgroup
}

\title{Multimodal Ambivalence/Hesitancy Recognition in Videos for Personalized Digital Health Interventions}

\renewcommand\footnotemark{}
\author{Manuela~González-González${^{3,4}}$, \; 
\textbf{Soufiane~Belharbi}${^1}$, \;
Muhammad~Osama~Zeeshan${^1}$, \; 
\textbf{Masoumeh~Sharafi}${^1}$, \;  \\
\textbf{Muhammad~Haseeb~Aslam}${^1}$, \;
\textbf{Lorenzo~Sia}${^1}$, \;
\textbf{Nicolas~Richet}${^1}$, \;
\textbf{Marco~Pedersoli}${^1}$, \;
\textbf{Alessandro~Lameiras~Koerich}${^2}$, \; \\ 
\textbf{Simon~L~Bacon}${^{3, 4}}$ \textbf{\&} \; 
\textbf{Eric~Granger}${^1}$\vspace{0.05in}\\
$^1$LIVIA, Dept. of Systems Engineering, ETS Montreal, Canada \vspace{0.02in}\\
$^2$LIVIA, Dept. of Software and IT Engineering, ETS Montreal, Canada \vspace{0.02in}\\
$^3$Dept. of Health, Kinesiology, \& Applied Physiology, Concordia University, Montreal, Canada \vspace{0.02in}\\
$^4$Montreal Behavioural Medicine Centre, CIUSSS Nord-de-l’Ile-de-Montréal, Canada\\
{\tt\scriptsize\{soufiane.belharbi, marco.pedersoli, alessandro.koerich, eric.granger\}@etsmtl.ca
}
\\
{\tt\scriptsize \{muhammad-osama.zeeshan.1,masoumeh.sharafi.1,muhammad-haseeb.aslam.1,lorenzo.sia.1,	nicolas.richet.1\}@ens.etsmtl.ca
}
\\
{\tt\scriptsize manuela.gonzalez@mail.concordia.ca, simon.bacon@concordia.ca
}
}

\newcommand{\ignore}[1]{}

\begin{document}
\doparttoc % Tell to minitoc to generate a toc for the parts
\faketableofcontents % Run a fake tableofcontents command for the partocs

\thepart{} % Start the document part
\parttoc % Insert the document TOC

\maketitle\thispagestyle{fancy}
\maketitle
% \lhead{\color{gray} \small \today}
\rhead{\color{gray} \small González et al. \;  [ACII 2026]}

\begin{abstract}
Using behavioural science, health interventions focus on behaviour change by providing a framework to help patients acquire and maintain healthy habits that improve medical outcomes. In-person interventions are costly and difficult to scale, especially in resource-limited regions. Digital health interventions offer a cost-effective approach, potentially supporting independent living and self-management. Automating such interventions, especially through machine learning, has recently gained considerable attention.
Ambivalence and hesitancy (A/H) play a primary role for individuals delaying, avoiding, or abandoning health interventions. A/H are subtle and conflicting emotions that place a person in a state between positive and negative evaluations of a behaviour, or between acceptance and refusal to engage in it. They manifest as affective inconsistency across modalities or within a modality, such as language, facial, vocal expressions, and body language.  While experts can be trained to recognize A/H, as done for in-person interactions, integrating them into digital health interventions is costly and less effective. Automatic A/H recognition is therefore critical for the personalization and cost-effectiveness of digital health interventions.
In this paper, we explore the application of deep learning models for A/H recognition in videos, a multi-modal task by nature. In particular, this paper covers three learning setups: supervised learning, unsupervised domain adaptation for personalization, and zero-shot inference via large language models (LLMs). Our experiments are conducted on the unique and recently published \bah video dataset for A/H recognition. Our results show limited performance, suggesting that more adapted multi-modal models are required for accurate A/H recognition. In addition, better methods for modeling spatio-temporal and multimodal fusion are necessary to leverage conflicts within and across modalities. 
Our code is publicly available:
\href{https://github.com/sbelharbi/ah-digital-health-interventions}{github.com/sbelharbi/ah-digital-health-interventions}\blfootnote{979-8-3195-2145-3/26/\$31.00 \copyright 2026 Crown.  DOI link (soon)}.
\end{abstract}


%============================== Introduction.

\begin{figure}[!htb]
\centering
  \centering
  \includegraphics[width=\linewidth]{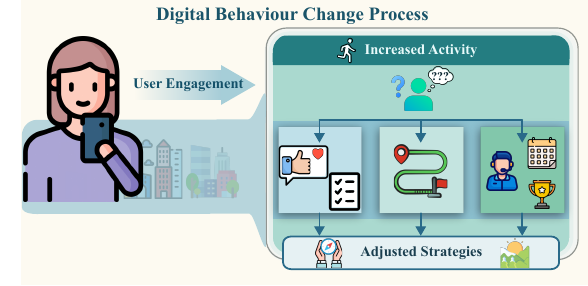}
  \caption{Conceptual illustration of the theoretical pathway by which ambivalence/hesitation (A/H) recognition can influence a digital behaviour change intervention: a user interacts with a smartphone app targeting physical activity, while the app detects A/H and adaptively adjusts cues and strategies to deliver more personalized support.}
  \label{fig:dhi}
  \vspace{-10pt}
\end{figure}

%%%%%%%%%%%%%%%%%%%%%%%%%%%%%%%%%%%%%%%%%%%%%%%%%%%%%%%%%%%%%%%%%%%%%%%%%%%%%%%%
\section{Introduction} 
\label{sec:intro}

The majority of medical health interventions aim to modify high-risk health behaviours that contribute to the growing burden of chronic diseases~\cite{heuveline22}. Such diseases account for substantial amounts of morbidity~\cite{buttorff17} and mortality~\cite{hacker24} worldwide~\cite{who25}. Major efforts have been put in place to positively modify these high-risk behaviours that impact human health and well-being through the development of effective behaviour change interventions~\cite{matthews24,parkinson25,Von-Klinggraeff24}. 
Despite these efforts, healthcare providers still struggle to support patients to acquire and maintain healthy behaviours~\cite{mulder25}. For instance, cardiovascular disease (CVD), which remains the main contributor to deaths before age 75~\cite{Nielsen17}, is largely caused by modifiable high-risk behaviours such as smoking, low-quality diet, physical inactivity, and medication non-adherence~\cite{olie24}. In addition, roughly 50\% of premature deaths from heart disease, stroke, and cancer are found to be related to high-risk behaviours ~\cite{mather15}. Though behaviour change-focused health interventions targeting high-risk behaviours have been shown to consistently improve health outcomes~\cite{Diabetes02}, they do not seem to work for everyone, suggesting that greater personalization is needed~\cite{hsu25,kaar17,zettler25}. 
While in-person health interventions are the most common form of intervention in healthcare systems~\cite{mccord22,santarossa18}, they are costly, have limited accessibility, and are difficult to scale. Digital health interventions are potentially cheaper, more scalable options, which could leverage common and affordable modalities such as web- and phone-based applications~\cite{widmer15}, intending to maintain the effectiveness of in-person interventions whilst also creating better access and a greater chance for personalization.

Achieving and maintaining long-term behaviour change includes overcoming ambivalence and hesitancy (A/H)~\cite{michie13,michie13move,voisard24}, which are the primary reasons for individuals to delay, avoid, or abandon health behaviour changes~\cite{conner08,conner02,manuel16,miller15,van24,williams24}. A/H is a subtle and conflicting emotion manifested by an affective inconsistency across multiple modalities or within a modality, such as facial and vocal expressions, and body language. The presence of this conflicting affect sets an individual in a state between positive and negative evaluations of a behaviour, or between acceptance and refusal to act, which often constitutes a barrier to initiating behaviour change and a trigger for discontinuing health interventions or behaviour change efforts. Healthcare providers (e.g., clinicians, therapists, interventionists) often identify A/H through a combination of speech and non-verbal cues, e.g., facial expressions and tone~\cite{heisel04,labbe22,miller15} during in-person interactions. However, human integration into digital health interventions and e-health tools can be costly and less effective. Therefore, designing robust automated methods for A/H recognition can provide a cost-effective alternative to adapt for individual users and operate in real-time and resource-limited environments. Leveraging machine learning (ML) for automating digital health interventions has recently gained much attention~\cite{bucher24,lisowska23,miller26}.

Recent research on ML in emotion recognition focuses mainly on seven basic discrete emotions in images or videos, e.g, `happy', `sad', and `surprised'~\cite{belharbi24-fer-aus,delazeri25,hanwei24,xue2022vision}. Other models in the literature predict ordinal levels, including levels linked to health states, e.g., pain, depression, and stress  estimation~\cite{aslam24,chaptoukaev23,sharafi26pers,zeeshan24,nasimzada24}. Other affects are continuous predictions such as valence-arousal~\cite{dong24,praveen23,praveen22,praveen21,teixeira21}.  However, real-world scenarios present more complex cases of emotions. Recently, there has been an increased interest in designing robust affect models for compound emotions, a case where a mixture of basic emotions is manifested~\cite{kollias23,Kollias25,richet-abaw-24}. In particular, compound emotions commonly occur in daily interactions. Recognizing affective states linked to compound emotions and health states is, however, more difficult to discern as they are subtle, ambiguous, and resemble basic emotions.  
A/H recognition is related to such tasks where intention and attitude are conflicting or in an intermediate (in-between) state, between willingness and resistance~\cite{macdonald15}, or positive and negative affect~\cite{armitage00}. This can manifest in how individuals express themselves and can be recognized~\cite{hayashi23} in their facial expressions, tone, verbal expressions, and body language.  
As a result, A/H is inherently multimodal, arising from subtle interactions among multiple cues both across modalities and within each modality. Unfortunately, such discord is extremely difficult to detect -- a task that requires human training.  This is a tedious and expensive procedure, leading to ineffective and less scalable digital interventions, especially under limited resources.
Integrating automated and reliable tools for A/H recognition can have a major impact on improving digital health interventions. Although A/H is a common topic in behavioural science~\cite{conner08,hohman16,manuel16}, it remains unexplored in the ML community, and as such, in the design of eHealth components.

In this paper, we explore the application of deep learning models for the newly introduced task, A/H recognition. Our experiments cover different learning scenarios, in particular supervised learning of A/H recognition at the frame and video level, where different modalities are explored, such as visual, audio, and text. We further extend our experiments to personalization through unsupervised domain adaptation. We also explore zero-shot inference using recent large language models (LLMs). To this end, we consider in our experiments the recent, unique, and video-based dataset for A/H recognition, the Behavioural Ambivalence/Hesitancy (\bah) dataset~\cite{gonzalez-26-bah}. 
Our results show that specialized multimodal and spatio-temporal models are required for more accurate A/H recognition. In addition, a new modality-fusion module able to detect conflicts within and across modalities is necessary.

%%%%%%%%%%%%%%%%%%%%%%%%%%%%
\section{Related Work} \label{sec:related-work}

This section provides an overview of research in affective computing related to behavioural science.

\subsection {Machine Learning for Affect Recognition}

\noindent\textbf{Basic Emotions.} An important line for ML research in affective computing is discrete facial emotion recognition (FER)~\cite{hanwei24,Kollias25,mao24posterpp,wang24}. This usually involves classifying facial images into one of seven or eight basic emotions, such as `Happy', `Sad', and `Surprised'.  Other works focus on videos~\cite{liu23facial,liu21identity,liu23} as well. There is also recent interest in designing robust FER methods that are interpretable~\cite{belharbi24-fer-aus,belharbi24ai,wang24five,xue2022vision}. They typically produce a heat map that indicates relevant image regions used by a model to perform a prediction. This is usually formulated as an attention map or a class activation map~\cite{murtaza25}. 
Other work aims to predict ordinal levels (i.e, ordered labels), including pain and stress estimation~\cite{aslam24,chaptoukaev23,zeeshan24,nasimzada24,sharafi26tta-cap,sharafi26pers}.
Dimensional recognition of emotions typically aims to estimate continuous valence and arousal values linked to emotions~\cite{dong24,praveen23,praveen22,praveen21}. 
Finally, another task related to emotion recognition is action units (AUs) detection~\cite{Luo22}. It aims to predict active AUs in the face.
Other works go further to estimate the intensity of AUs~\cite{fan20}, or both~\cite{Sanchez18}, which is a much more challenging task.

\noindent\textbf{Compound Emotions.} Real-world scenarios often present complex emotions that combine basic ones.  There has been recent interest in building affective computing models to predict compound emotions, a case where a mixture of basic emotions is expressed~\cite{kollias23,Kollias25,richet-abaw-24}. These are shown in several practical real-world applications, as such complex emotions occur in daily interactions. However, they are more difficult to recognize as they are subtle, ambiguous, and resemble basic emotions. 
A/H recognition is related to a compound emotion recognition task where intention and attitude are conflicted or in an in-between state, between willingness and resistance~\cite{macdonald15}, or positive and negative affect~\cite{armitage00,conner02}. This can manifest in how individuals express themselves and can be recognized~\cite{hayashi23} in their facial expressions, tone, verbal, and body language. As a result, A/H exhibits a multimodal nature that arises from the subtle interconnection between different cues. Unfortunately, such dissonance is extremely difficult to detect, a task that requires human training. This is a tedious and expensive procedure~\cite{mantena25}, leading to ineffective and less scalable digital health interventions under limited resources. Assisting healthcare providers with automatic, reliable, and inconspicuous tools to help them recognize A/H can have a major impact on improving digital health interventions.

\subsection{Behavioural Science}
\noindent\textbf{Health-Related Behaviour Change and Non-Communicable Diseases.}
High-risk health behaviours, such as tobacco use, physical inactivity, low-quality diet, and harmful alcohol consumption~\cite{olie24}, are responsible for the vast majority of non-communicable diseases, which include cardiovascular disease, type 2 diabetes, cancer, and chronic respiratory illnesses. According to the World Health Organization~\cite{ortiz25, heuveline22}, non-communicable diseases account for approximately 74\% of global deaths, and these outcomes are disproportionately influenced by modifiable behavioural factors~\cite{mather15,Nielsen17}. Consequently, health-related behaviour change has become a primary target for preventive and therapeutic interventions~\cite{Diabetes02}. Traditional interventions rely on in-person clinical interactions and delivery, which remain foundational to behavioural health practice~\cite{odonnel19}. These interactions provide opportunities for clinicians to detect A/H and other complex affective states, often through subtle verbal and nonverbal cues~\cite{hall95}. Despite the growing shift toward digital platforms and interventions, face-to-face interactions remain the gold standard for identifying emotional and cognitive responses, leading to insights that are essential for tailoring behaviour change strategies~\cite{hsu25, kaar17, zettler25}. Efforts to change health behaviours over the long term are complex~\cite{middleton13}. Individuals often experience A/H, understood as fluctuating between intention and resistance, when attempting to adopt healthier habits~\cite{armitage00, conner02}. In traditional healthcare contexts, providers rely on both verbal communication and non-verbal cues (e.g., tone, gestures, facial expressions) to recognize and address such conflicts. This in-person interaction allows for nuanced support that can adapt to a patient’s readiness for change~\cite{davidson20}. The purpose of developing multimodal A/H recognition systems is to capture and replicate this nuanced understanding of the patient behaviour change process within digital health interventions, thereby supporting clinicians and scaling behavioural health care through digital health interventions.

\noindent\textbf{Multimodal Cues and the Detection of Complex Emotions.}
Identifying complex emotional states such as ambivalence, resistance, or hesitancy is crucial for tailoring behavioural health interventions. Research in psychology and human-computer interaction has shown that complex emotional states, such as ambivalence, uncertainty, or defensiveness, are communicated through a combination of facial expressions, body posture, vocal tone, speech patterns, and physiological responses~\cite{guo18}. In digital contexts, however, the absence of physical presence makes this task more difficult. Recent research in psychology and computer science has focused on the use of multimodal cues, such as facial expressions, voice tone, body posture, and physiological responses, as proxies for emotional and motivational states~\cite{kraack24,yan24}. These cues can reveal underlying emotional conflict or uncertainty that might not be captured by self-report alone. Combining multiple input channels (e.g., audio-visual data) has been shown to enhance the accuracy of emotion recognition systems. For instance, multimodal datasets are being used to train models that detect affective states like confusion, frustration, and mixed emotions, which are highly relevant in contexts such as education, mental health, and health-related behaviour change. By incorporating these data streams, researchers can better approximate the nuanced human capacity for reading emotions, paving the way for emotionally aware systems~\cite{he20,zhao21}. 

\noindent\textbf{Affective Computing and Personalized Digital Health Interventions. } ,
Affective computing, a subfield of artificial intelligence (AI) focused on recognizing, interpreting, and responding to human emotions and holds promise for advancing personalized digital health interventions. By leveraging emotion-aware algorithms, digital platforms can better understand users' psychological readiness and tailor support accordingly~\cite{lokhande24,vairamani24}. For example, interventions that dynamically respond to detected signs of resistance or disengagement may improve user retention and behavioural outcomes~\cite{xu20}. Incorporating affective computing into digital health technologies also allows dynamic tailoring of content based on users’ real-time affect, responsive dialogue, mimicking the adaptability found in face-to-face interactions~\cite{yardley16}. Recent advancements in conversational agents, voice analysis, and facial expression recognition have made it possible for digital interventions to adapt content delivery based on real-time emotional assessments~\cite{khanna22}. This not only improves user engagement but also enhances intervention effectiveness by ensuring messages are delivered in an emotionally congruent and contextually appropriate manner~\cite{hornstein23}.

%%%%%%%%%%%%%%%%%%%%%%%%
\section{\bah Dataset} \label{sec:dataset}

\begin{figure}[!t]
\vspace{-20pt}
\centering
  \includegraphics[width=\linewidth]{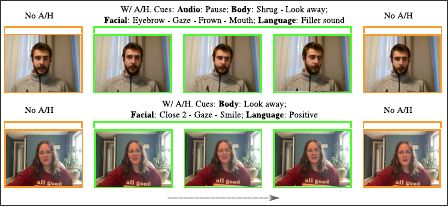}
  \caption{\bah examples of frames with (green) and without (orange) A/H with cues detailed in~\cite{gonzalez-26-bah}.}
  \label{fig:ah-cues-body-language}
\end{figure}

For all our experiments, we use the public \bah dataset~\cite{gonzalez-26-bah,gonzalez-26-ident} (Fig.~\ref{fig:ah-cues-body-language}) for A/H recognition in videos. It is composed of videos from 300 participants. They were recorded in their own environments, answering seven questions, some of which were designed to elicit A/H. The dataset provides three modalities: visual, audio, and transcript (text). It is annotated at frame- and video-level. It contains a total of 1,427 videos and 10.60 hours of recording. We use the standard data partitions originally provided by the authors~\cite{gonzalez-26-bah}: train set (195 participants: 778 videos, 501,970 frames), validation set (30 participants: 124 videos, 79,538 frames), and test set (the remaining 75 participants: 525 videos, 335,110 frames). For evaluation, we use the same metrics in~\cite{gonzalez-26-bah}: (i) Average \fonescore score (\avgfonescore), which is the unweighted mean of \fonescore of the positive and negative classes; (ii) Average precision score (\apscore, threshold list between 0 and 1, with a step of 0.001) of the positive class, which accounts for performance sensitivity to the model's confidence; 
(iii) In addition, we include weighted average recall (\war), which measures the overall classification accuracy.
The following sections provide experiments over three learning scenarios: supervised learning, personalization using domain adaptation, and zero-shot inference using MLLMs. In all our experiments, we use the following references: ZS: Zero-shot. FT: Fine-tuning. NA: Not available.

% =======================================================================
%                  SUPERVISED LEARNING
% =======================================================================

\section{Supervised Learning Experiments} 
\label{sec:sup-learning}

The performance of models is compared for supervised frame- and video-level classification. Following the \bah annotation, a 2-class classification task is considered where the positive class indicates the presence of A/H, while the negative class models the absence. All three learning scenarios follow this same setup. 
We explore here the following aspects: the impact of single vs multimodal learning for frame-level classification; the impact of individual modalities/multimodal/fusion, temporal-modeling/context; and video-level classification.

\subsection{Pre-processing of Modalities}
\label{subsec:preproces-modalities}

\noindent\textbf{Visual}. Frames from each video captured at 24 fps are extracted, and for each frame, faces are cropped and aligned using the RetinaFace model~\cite{deng19retinaface}. The face with the highest score is stored in case multiple faces are detected in a frame. Faces are resized to 256$\times$256 and stored as RGB images with filenames that maintain the order of frames.

\noindent\textbf{Audio}. We follow a standard procedure to process audio data~\cite{praveen22,richet-abaw-24,zhang23}. For the audio modality, we first convert videos to single audio channels (mono) with a 16 kHz sampling rate into WAV format. The log Melspectrogram features are extracted using the VGGish model~\cite{hershey17}\footnote{\href{https://github.com/harritaylor/torchvggish}{github.com/harritaylor/torchvggish}}. A hop of 1/FPS of the raw video is used to extract the spectrograms to synchronize audio with other modalities.

\noindent\textbf{Text}. \bah dataset captures the audio of participants. We consider audio transcripts as an extra modality that can help in recognizing A/H since text is a significant cue used by annotators~\cite{gonzalez-26-bah}. To this end, we transcribe the audio of each video and detect the language using the Whisper model~\cite{radford23} \footnote{Whisper large-v3 multilingual: \href{https://huggingface.co/openai/whisper-large-v3}{huggingface.co/openai/whisper-large-v3}}. 
Word-level features are then extracted using the BERT Base Uncased model~\cite{Devlin19}\footnote{\href{https://pypi.org/project/pytorch-pretrained-bert/}{pypi.org/project/pytorch-pretrained-bert/}}. A word may span more than one frame. To synchronize with other modalities, a word-level feature is repeated per its timestamp for all the frames that correspond to the word.

%%%
\subsection{Pre-training of Visual Backbone}
\label{subsec:visual-pretrain-backbones}
For audio and text modality, features are extracted offline and stored as described above. For visual modality, we explore different architectures, including CNN- and ViT-based~\cite{Dosovitskiy21}. For the ResNet family, we consider ResNet18, 34, 50, 101, and 152~\cite{heZRS16}, and for the ViT family, we consider a recent model designed for basic emotion recognition, APViT~\cite{xue2022vision}. First, each model is pre-trained on a basic emotion recognition task, including these emotions: ``anger'', ``disgust'', ``fear'', ``happiness'', ``sadness'', ``surprise''. To this end, we used a large mixed dataset composed of 3 public common datasets for emotion recognition using images: \rafdb~\cite{li17}, \affectnet~\cite{MollahosseiniHM19}, and \affwildtwo~\cite{kollias19}. This amounts to more than $\sim$0.54 million training images. Models are trained for basic emotion classification for 60 epochs with a batch size of 1,424 samples. 
Standard cross-entropy loss and stochastic gradient descent (SGD) are used for training. Once pretrained, each model is further fine-tuned on the \bah train set for A/H recognition. To account for class imbalance, we perform under-sampling of the negative class over the training set. This is achieved by randomly sampling the negative class samples to match the number of positive class samples. 
The backbones of each model are used later for feature extraction of the visual modality.

{
\setlength{\tabcolsep}{3pt}
\renewcommand{\arraystretch}{1.1}
\begin{table}[!t]
\centering
\resizebox{\linewidth}{!}{%
\centering
%\small
\color{black}
\setlength{\tabcolsep}{2pt}
\begin{tabular}{lccccc ccc}
\hline
% &&  \textbf{Tiles}    &  &   &        &&  \textbf{Patches}   &  &   & \\ \hline \hline
% %\cline{1-6} \cline{8-11}\\
              && \multicolumn{2}{c}{Without context} && \multicolumn{2}{c}{With context (TCN)} \\ \hline
Backbone  && \avgfonescore  & \apscore &&  \avgfonescore & \apscore \\ \hline 
%\cline{1-1}\cline{3-6} \cline{8-11}\\
% \hline
APViT~\cite{xue2022vision} {\fontsize{7}{12} \selectfont (TAFFC'22)}  &&   0.5051   & 0.1906 && 0.5019 & 0.2069\\ 
ResNet18~\cite{heZRS16}    {\fontsize{7}{12} \selectfont (CVPR'16)}   &&  0.5074 & 0.1940    && 0.5079 & 0.1993\\ 
ResNet34~\cite{heZRS16}    {\fontsize{7}{12} \selectfont (CVPR'16)}   &&  \textbf{0.5138} & 0.1952    && 0.4998 & 0.1984\\ 
ResNet50~\cite{heZRS16}    {\fontsize{7}{12} \selectfont (CVPR'16)}   &&  0.4737 & 0.1942    && 0.4985 & 0.1915\\ 
ResNet101~\cite{heZRS16}   {\fontsize{7}{12} \selectfont (CVPR'16)}   &&  0.4929 & \textbf{0.1967}    && \textbf{0.5165} & \textbf{0.2070}\\ 
ResNet152~\cite{heZRS16}   {\fontsize{7}{12} \selectfont (CVPR'16)}   &&  0.4889 & 0.1843    && 0.5084 & 0.2058\\ 
\hline
\end{tabular}
}
\caption{Impact of architecture and context: Performance of the visual modality on the test subset of \bah for frame-level classification.}
\label{tab:context-perf-frame}
\vspace{-1em}
\end{table}
}

\subsection{Importance of Contextual Learning}  
\label{subsec:context-results}
In this section, we investigate whether context modeling can improve A/H recognition. This is particularly interesting since A/H does not occur instantly but within a context. The text and audio modalities already capture context in their features, making them less efficient in answering this question. However, we can obtain frame features with and without any context. Therefore, we consider the visual modality with different backbones to answer our question. 

\begin{figure}[!th]
\centering
\includegraphics[width=\linewidth]{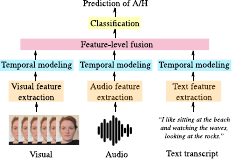} 
\caption{Multimodal model used for baseline evaluation~\cite{richet-abaw-24}.}
\label{fig:multimodal-model}
\vspace{-10pt}
\end{figure}

In the case without context, models simply train on independent frames without considering any context or dependency between them. Inference is done in the same way. In the case of using context, both training and inference leverage temporal dependency between frames. To this end, a window of adjacent frames is fed to the model. We then use a temporal convolutional network (TCN)~\cite{bai18} after the visual backbone to capture relations between frame embeddings. The window length defines the extent of the context. 

\begin{figure*}[!ht]
     \centering
     \begin{subfigure}[b]{0.49\textwidth}
         \centering
         \includegraphics[width=\textwidth]{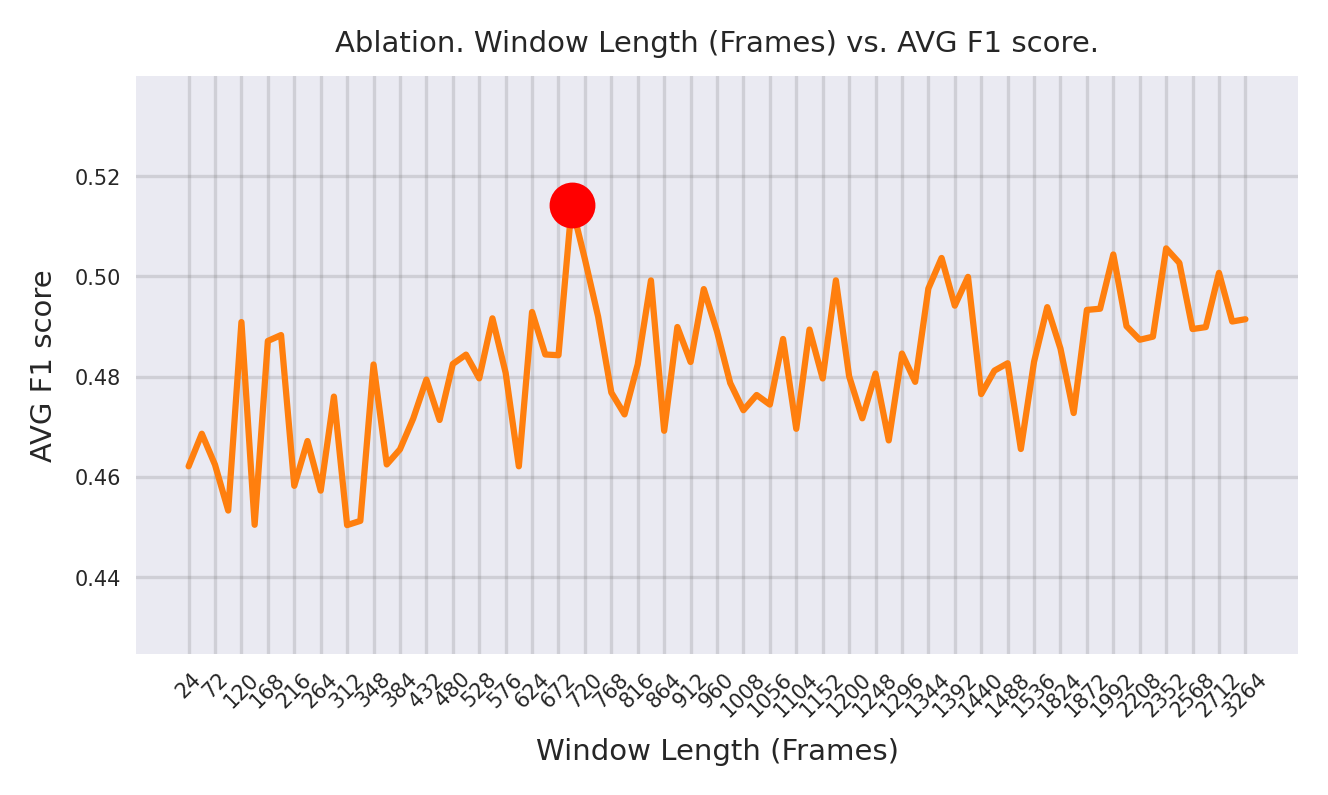}
     \end{subfigure}
     \begin{subfigure}[b]{0.49\textwidth}
         \centering
         \includegraphics[width=\textwidth]{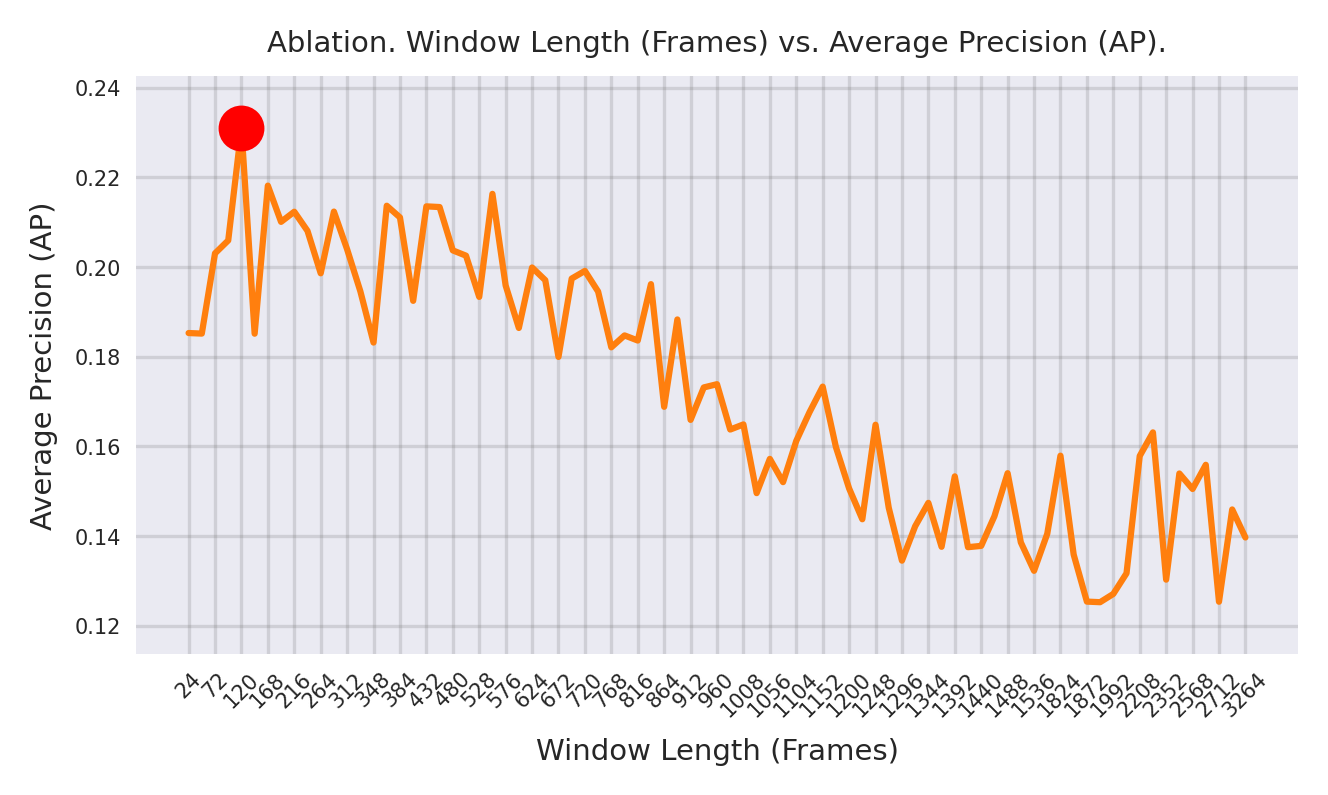}
     \end{subfigure}
        \caption{Impact of window length on the performance of frame-level classification when using visual modality alone (ResNet18): \avgfonescore, and \apscore. Best performance is indicated in red dot.}
        \label{fig:ablation-context}
\end{figure*}

First, we conduct an ablation to study the impact of the window length on the performance of recognizing A/H. To this end, we use a window length from 24 to 3264 with a step of 1 second (24 frames). Figure~\ref{fig:ablation-context} shows the obtained results. By considering \wfonescore, performance improves with the increase of the context where it can reach above 0.825. On the other hand, \apscore prefers small context. Using a small context of few seconds could be a good compromise for all the metrics.

Regardless of the context, Table \ref{tab:context-perf-frame} shows low accuracy over \apscore, highlighting the difficulty of recognizing A/H based on images alone. In particular, \apscore is below 0.2070. On the other hand, \avgfonescore reaches 0.5138. 
Overall, using context improves the performance of all metrics across all architectures. This is expected, as A/H does not usually occur at a single frame but within a context. This makes its recognition from a single frame challenging. We recall that the average length of A/H segments in \bah spans 103 frames (or $\sim$4.29 sec). Future work should account for the temporal context more efficiently to achieve better performance. 
Since the ResNet101 architecture yields the best performance, it is used in the rest of our experiments.

{
\renewcommand{\arraystretch}{1.1}
\begin{table}[ht!]
\centering
\resizebox{\linewidth}{!}{%
\centering
%\small
\setlength{\tabcolsep}{2pt}
\begin{tabular}{lccccccc}
\hline
& \multicolumn{7}{c}{Modalities}\\
\cline{2-8}
& V & A & T & V+A & V+T & A+T & V+A+T\\
\cline{2-8}
\avgfonescore & 0.5165 & 0.4658 & 0.5497 & 0.5205 & 0.5547 & \textbf{0.5586} & 0.5502\\
\apscore & 0.2070 & 0.2238 & 0.2519  & 0.2225 & 0.2479 & \textbf{0.2609} & 0.2548\\
\hline 
\end{tabular}
}
\caption{Frame-level classification performance of multimodal models on the \bah dataset.{\fontsize{8}{12} \selectfont V: Video, A: Audio, T: Text}.
}
\label{tab:multimodal-perf-frame}
\vspace{-10pt}
\end{table}
}

\subsection{Multimodal Baselines}
\label{subsec:multimodal-results}

Since A/H has a multimodal nature, we explore the impact of using different modalities, including visual, audio, and transcript (text), using the architecture shown in Fig.~\ref{fig:multimodal-model}~\cite{richet-abaw-24}. Results are reported in Table~\ref{tab:multimodal-perf-frame}. 
Using the text modality alone yields better performance, \apscore of 0.2519 and \avgfonescore of 0.5497, compared to visual or text modalities over both metrics. 
Combining a pair of modalities improves the performance to 0.5586 for \avgfonescore, and 0.2609 for \apscore, in the case of audio and text. Combining the three modalities slightly reduces performance in terms of \apscore, suggesting that more adapted fusion techniques are needed to recognize affect conflicts between/within modalities.

Table~\ref{tab:fusion-perf-frame} shows the impact of using different feature fusion techniques, including simple concatenation (CAN)~\cite{zhang23}, co-attention (LFAN)~\cite{zhang23}, transformer-based fusion (MT)~\cite{waligora-abaw-24}, and cross-attention fusion (JMT)~\cite{waligora-abaw-24}. We observe that the way of leveraging the interaction between the three modalities is a key factor. LFAN and CAN fusion lead over both metrics. Future work should pursue more adapted methods to A/H. Since A/H are usually expressed as a conflict between willingness and resistance, they can be perceived through a parallel affect conflict between modalities and/or within modalities. 
For instance, a participant could say a sentence to convey a meaning, but their facial expression, body behaviour, or tone may carry a contradictory emotion. Understanding such subtleties and interconnections between different cues in different modalities could play an important role in designing robust methods for A/H recognition in videos.

{
\setlength{\tabcolsep}{3pt}
\renewcommand{\arraystretch}{1.1}
\begin{table}[hbt!]
\centering
\resizebox{\linewidth}{!}{%
\centering
%\small
\color{black}
\begin{tabular}{lcccccccc}
\hline
% &&  \textbf{Tiles}    &  &   &        &&  \textbf{Patches}   &  &   & \\ \hline \hline
% %\cline{1-6} \cline{8-11}\\
Method && Fusion Approach&& \avgfonescore & \apscore \\ \hline 
%\cline{1-1}\cline{3-6} \cline{8-11}\\
LFAN~\cite{zhang23} {\fontsize{7}{12} \selectfont (CVPRW'23)}     && Co-attention    &&  0.5502   & 0.2548    \\ 
CAN~\cite{zhang23} {\fontsize{7}{12} \selectfont (CVPRW'23)}  && Concatenation &&  \textbf{0.5526} & \textbf{0.2631}    \\ 
MT~\cite{waligora-abaw-24}  {\fontsize{7}{12} \selectfont (CVPRW'24)} && Transformer &&  0.5137  & 0.2134    \\ 
JMT~\cite{waligora-abaw-24} {\fontsize{7}{12} \selectfont (CVPRW'24)} && Cross-attention &&  0.5241  & 0.2139    \\
\hline
\end{tabular}
}
\caption{Impact of feature fusion methods on test set of the \bah dataset at frame-level classification.}
\label{tab:fusion-perf-frame}
\vspace{-10pt}
\end{table}
}

\subsection{Video-level classification}
\label{subsec:video-classif}

We conducted experiments on the Video-FocalNet~\cite{wasim23videofocalnet} model, specialized in video-based classification. It can be trained to directly predict a video class based on sampled frames. However, it accounts only for the visual modality. We used cropped-and-aligned faces as input. We experimented with different architectures: tiny, small, and base. Pretrained weights provided by~\cite{wasim23videofocalnet} over Kinetics-400 dataset~\cite{kay17kinetics400} are used for initialization. We also compare with EmoCLIP~\cite{foteinopoulou24emoclip}, which is a CLIP-based video classification model. We use the pretrained emotion model from~\cite{foteinopoulou24emoclip} for the Zero-Shot setting, and we also train a model following the same architecture with weights initialized using CLIP (ViT-B/32) for A/H recognition. We follow the methodology from~\cite{foteinopoulou24emoclip} to generate class-level descriptions. Following~\cite{wasim23videofocalnet}, we sampled 8 frames per video. The stride between the frames depends on the video length. This is done during training and testing. We used a batch size of 64 (videos) for all models. The models are fine-tuned over the \bah dataset for 60 epochs. Table~\ref{tab:context-perf-video} shows that basic emotion understanding of EmoCLIP does not directly translate to ambivalence recognition. Training improves performance, but a specialized model yields better performance on average over both metrics. Although performance fluctuates between the three architectures, Video-FocalNet base-case yields an \apscore of ${0.6732}$ and \avgfonescore of ${0.5663}$, leading the board.

{
\setlength{\tabcolsep}{3pt}
\renewcommand{\arraystretch}{1.1}
\begin{table}[!t]
\centering
\resizebox{\linewidth}{!}{%
\centering
%\small
\color{black}
\begin{tabular}{lcc cccc}
\hline
% &&  \textbf{Tiles}    &  &   &        &&  \textbf{Patches}   &  &   & \\ \hline \hline
% %\cline{1-6} \cline{8-11}\\
              % && \multicolumn{2}{c}{Without context} && \multicolumn{2}{c}{With context (TCN)} \\ \hline
Backbone  && Setting & \avgfonescore & \apscore \\ \hline 
%\cline{1-1}\cline{3-6} \cline{8-11}\\
% \hline
EmoCLIP~\cite{foteinopoulou24emoclip}  {\fontsize{7}{12} \selectfont (FG'24)}                && ZS & 0.4525 & 0.5894\\
EmoCLIP~\cite{foteinopoulou24emoclip}  {\fontsize{7}{12} \selectfont (FG'24)}                && FT & 0.5222 & 0.6275 \\ 
Video-FocalNet (tiny)~\cite{wasim23videofocalnet}   {\fontsize{7}{12} \selectfont (ICCV'23)} && FT & 0.5294 & 0.6545\\
Video-FocalNet (small)~\cite{wasim23videofocalnet}  {\fontsize{7}{12} \selectfont (ICCV'23)} && FT & 0.5333 & 0.6558\\
Video-FocalNet (base)~\cite{wasim23videofocalnet}   {\fontsize{7}{12} \selectfont (ICCV'23)} && FT & \textbf{0.5663} & \textbf{0.6732}\\
\hline
\end{tabular}
}
\caption{Video classification performance on \bah test set.}
\label{tab:context-perf-video}
% \vspace{-1em}
\end{table}
}

% =======================================================================
%                  SFDA
% =======================================================================

\section{Personalization using Domain Adaptation}
\label{sec:personalization}

Domain adaptation (DA)~\cite{han2020personalized,li2018deep} is a promising approach for personalized expression recognition, where a model trained on labeled source data is adapted to target domain data corresponding to individual users. Recent work has emphasized subject-based domain adaptation~\cite{sharafi26pers,sharafi2025disentangled,zeeshan24,zeeshan2025progressive}, in which each subject is a distinct domain. In this paper, each participant in the test set is considered as an independent target domain for the task of frame-level classification.

We follow the standard personalization protocol used in prior work~\cite{zeeshan24,sharafi2025disentangled}, where each target subject is split into train, validation, and test sets. Because the \bah dataset is imbalanced, each split is constructed to preserve a balanced representation of positive and negative samples.
We establish the following baseline methods: \textbf{Source-only}: the source-trained model is directly tested on the target subject without adaptation.
\textbf{Unsupervised Domain Adaptation (UDA)}: the model is adapted using labeled source data and unlabeled target data. We consider: (i) MMD~\cite{sejdinovic2013equivalence}, which reduces the source--target discrepancy, and (ii) ACPL~\cite{zeeshan24}, a subject-based method that combines pseudo-labeling and MMD-based alignment.
\textbf{Source-Free (Unsupervised) Domain Adaptation (SFDA)}: It relies on unlabeled target data without access to source samples. We consider: (i) SHOT~\cite{liang2020we}, which combines information maximization with prototype-guided pseudo-labeling, and (ii) NRC~\cite{yang2021exploiting}, which enforces neighbourhood consistency through reciprocal nearest neighbours and affinity-weighted self-regularization.
\textbf{Test-Time Adaptation (TTA)}: the model is adapted at inference time using only unlabeled target samples, without a separate adaptation stage or access to source data at deployment. TTA is especially attractive for practical personalization. We evaluate several recent TTA methods, including CLIP-based personalized adaptation approaches. Performance is reported using \avgfonescore and weighted average recall (\war).
\textbf{Oracle}: the model is fine-tuned using labeled target-subject data, providing an upper bound. For all experiments, we use a ViT-based visual backbone.

{
\renewcommand{\arraystretch}{1.1}
\begin{table}[t!]
\centering
\resizebox{\linewidth}{!}{%
\centering
%\small
\color{black}
\setlength{\tabcolsep}{1.5pt}
\begin{tabular}{llccc}
\hline
% &&  \textbf{Tiles}    &  &   &        &&  \textbf{Patches}   &  &   & \\ \hline \hline
% %\cline{1-6} \cline{8-11}\\
& Methods  && \avgfonescore & \apscore \\ \hline 
%\cline{1-1}\cline{3-6} \cline{8-11}\\
Source-only   &      &&  $0.4894 \pm  0.0999$    &  $0.3565 \pm 0.1841$   \\\cline{1-2}
%\multicolumn{4}{c}{UDA}\\\hline

\multirow{2}{*}{UDA} & 
MMD~\cite{sejdinovic2013equivalence}      {\fontsize{7}{12} \selectfont (Ann.Stat.'13)}    &&  $0.4931\pm0.0943$    &  $0.3589\pm0.1831$   \\ 
& ACPL~\cite{zeeshan24}  {\fontsize{7}{12} \selectfont (FG'24)} &&   $\mathbf{0.5417\pm0.0728}$  &  $\mathbf{0.3739\pm0.1789}$   \\ \cline{1-2}

\multirow{3}{*}{SFDA} & 
 SHOT~\cite{liang2020we} {\fontsize{7}{12} \selectfont (ICML'20)} && $0.4919\pm0.1056$   &  $0.3520\pm0.1656$ \\
& NRC~\cite{yang2021exploiting} {\fontsize{7}{12} \selectfont (NeurIPS'21)} &&      $\mathbf{0.5174 \pm 0.1041}$   & $\mathbf{0.3688\pm0.1487}$     \\ 
& Oracle &&   $0.5864\pm0.0751$   &  $0.4181\pm0.1750$   \\\hline

\multirow{8}{*}{TTA} & 
     && \avgfonescore & \war \\ \cline{3-5}
& TPT~\cite{shu2022test} {\fontsize{7}{12} \selectfont (NeurIPS'22)} && $0.3970 \pm 0.0485$ & $0.6568 \pm 0.1076$ \\
& TDA~\cite{karmanov2024efficient} {\fontsize{7}{12} \selectfont (CVPR'24)} && $0.3990 \pm 0.0419$ & $0.6522 \pm 0.1092$ \\
& DPE~\cite{zhang2024dual} {\fontsize{7}{12} \selectfont (NeurIPS'24)} && $0.3940 \pm 0.0510$ & $0.6676 \pm 0.1135$ \\
& PromptAlign~\cite{hassan2023align} {\fontsize{7}{12} \selectfont (NeurIPS'23)} && $0.3970 \pm 0.0785$ & $0.6720 \pm 0.1125$ \\
& ReTA~\cite{liang2025advancing} {\fontsize{7}{12} \selectfont (ACM MM'25)} && $0.3980 \pm 0.0833$ & $0.6763 \pm 0.1081$ \\
& T3AL~\cite{liberatori2024test} {\fontsize{7}{12} \selectfont (CVPR'24)} && $0.4070 \pm 0.0400$ & $0.6794 \pm 0.1156$ \\
& TTA-CaP~\cite{sharafi26tta-cap} {\fontsize{7}{12} \selectfont (CoRR'26)} && $\mathbf{0.4112 \pm 0.0455}$ & $0.6929 \pm 0.1128$ \\
& CLIP-AUTT~\cite{zeeshan26clipautt} {\fontsize{7}{12} \selectfont (CoRR'26)} && $0.4111 \pm 0.0446$ & $\mathbf{0.6983 \pm 0.1391}$\\\hline
\end{tabular}
}
\caption{Performance of UDA, SFDA, and TTA settings, with Source-only and Oracle as reference points. Results are reported as mean $\pm$ standard deviation over all target subjects.}
\label{tab:persona_uda}
\vspace{-10pt}
\end{table}
}

\noindent\textbf{Results and Analysis. }
The average performance across target participants is reported in Table~\ref{tab:persona_uda}. Overall, domain adaptation improves personalized A/H detection over the Source-only baseline. In the UDA setting, ACPL~\cite{zeeshan24} achieves the best performance, with an \avgfonescore of 0.5417 and an \apscore of 0.3739, clearly outperforming MMD~\cite{sejdinovic2013equivalence}. This highlights the benefit of combining pseudo-labeling with subject-aware alignment. By contrast, MMD~\cite{sejdinovic2013equivalence} and SHOT~\cite{liang2020we} yield only limited gains, suggesting that simple alignment or source-free hypothesis transfer is not sufficient to handle the strong subject shift in \bah. Among SFDA methods, NRC~\cite{yang2021exploiting} outperforms SHOT~\cite{liang2020we} and improves over Source-only by about 3\% in \avgfonescore and 1.2\% in \apscore, showing the value of exploiting local neighbourhood structure in the target domain. Still, both SFDA methods remain below ACPL~\cite{zeeshan24}, indicating the advantage of source-data access during adaptation. The Oracle result remains substantially higher than all unsupervised methods, confirming that there is still room for improvement in personalized positive-class detection. For TTA, all methods improve \war over Source-only, indicating better subject adaptation at inference time. However, gains in \avgfonescore are limited. TTA-CaP~\cite{sharafi26tta-cap} achieves the best \avgfonescore (0.4112), while CLIP-AUTT~\cite{zeeshan26clipautt} obtains the best \war (0.6983). This suggests that TTA improves overall adaptation and decision consistency, but it is less effective than UDA for handling class imbalance. Its source-free and deployment-friendly setting nevertheless makes it attractive for practical personalization.

% =======================================================================
%                  ZERO-SHOT INFERENCE
% =======================================================================
\section{Zero-shot Inference: Multimodal Large Language Models}
Multimodal LLMs (MLLMs) have gained significant attention in the affective computing space due to their ability to infer cross-modal dynamics across the visual, audio, and textual modalities. The problem of detecting A/H in videos is inherently multimodal as it also requires capturing the cross-modal inconsistency. To get out-of-the-box performance of existing SOTA MLLMs, we perform zero-shot inference using the Video-LLaVA-7B-hf \cite{vid-llava} using visual (full frame) and transcript modalities. Since the performance of an MLLM or LLMs in general can be heavily influenced by the query prompt, we experiment with different variations of the prompts. Table ~\ref{tab:prompt-summary} summarizes the different prompt variations used for zero-shot inference.    
We also experiment with AffectGPT~\cite{lian25}, a new and specialized MLLM for affective computing. It leverages audio and cropped faces in addition to text prompts.

{
\setlength{\tabcolsep}{3pt}
\renewcommand{\arraystretch}{1.1}
\begin{table}[ht!]
\centering
\resizebox{\linewidth}{!}{%
\centering
%\small
\color{black}
\begin{tabular}{c|l}
 \hline
& \multicolumn{1}{|c}{\textbf{Prompt}}                \\ \hline
Simple           & \begin{tabular}[c]{@{}l@{}}`Classify the emotion in the video as either \textit{Non-Ambivalent} \\ or \textit{Ambivalent}.'  Respond with only one word: \end{tabular}                             \\ \hline
Definition 1     & \begin{tabular}[c]{@{}l@{}}`Definition: Ambivalence is the state of having contradictory or \\ conflicting feelings or attitudes towards something or someone \\ simultaneously. Classify the emotion in the video as either \\\textit{Non-Ambivalent} or \textit{Ambivalent}.’ Respond with only one word: \end{tabular}                               \\ \hline
Definition 2     & \begin{tabular}[c]{@{}l@{}} `Definition: Ambivalence and   Hesitancy is understood as the \\ simultaneous experience of desires for change and against change. \\ Classify the emotion in the video as either   \textit{Non-Ambivalent} or \\ \textit{Ambivalent}.’  Respond with only one word: \end{tabular}                                                   \\ \hline
Transcript + Def 1 & \begin{tabular}[c]{@{}l@{}}`Video transcript: \textcolor{red}{\{transcript\}}. Definition: Ambivalence is the  \\ state of having contradictory or conflicting feelings or attitudes \\ towards something or someone simultaneously. Classify the \\ emotion in the video as either \textit{Non-Ambivalent} or  \textit{Ambivalent}.' \\ Respond with only one word: \end{tabular} \\ \hline
Transcript + Def 2 & \begin{tabular}[c]{@{}l@{}}`Video transcript: \textcolor{red}{\{transcript\}}. Definition: Ambivalence and \\Hesitancy understood as the simultaneous experience of desires \\ for change and against change. Classify the emotion in the video \\ as either \textit{Non-Ambivalent} or  \textit{Ambivalent}.' \\Respond with only one word:\end{tabular}                    \\ \hline
\end{tabular}
}
\caption{Prompt variations used for zero-shot inference.}
\label{tab:prompt-summary}
\end{table}
}

For \emph{frame-level} prediction, we adopt a segment-wise strategy, where the entire video is divided into 8-frame chunks and passed through the model using a sliding window. This way, the model sees all the frames in each video. The segment prediction is then duplicated for each frame within. 
The model's output, 'non-ambivalent' or 'ambivalent', is mapped to 0 and 1, respectively, to match the ground truth.

\noindent\textbf{Results and Analysis. }
Table~\ref{tab:mllm-video-frame} shows the results obtained for frame-level predictions using different prompts. The best results for frame-level prediction are obtained using the 'Transcript + Def 1' prompt, where the actual transcript of the video is also provided, along with a straightforward definition of A/H. The model performs better with a more straightforward definition of the concept of ambivalence, with Definition 1.

For \emph{video-level} prediction, the full video is used, and the transcript is embedded in the prompt. The model selects 8 uniformly spaced frames from the video and predicts a single output. Similar to frame-level predictions, the model's output is mapped to 0 and 1, and the performance metrics are calculated. Table~\ref{tab:mllm-video-frame} presents the video-level prediction results. Similar to frame-level predictions, the 'simple' prompt, without any context on the definition or the transcript, performs the worst and predicts all samples as 'Non-Ambivalent'. A similar trend is also observed here -- adding the definition and the transcript substantially affects the model performance. The last form of the table shows the results with fine-tuning. The improvement from 0.6341 to 0.7142 shows that the task of A/H detection is complex and that simple fine-tuning yields limited gains. Special modeling is required to capture all the different cues important in A/H detection. The AffectGPT model~\cite{lian25} achieves comparable performance but shows reduced sensitivity to prompt variations.

{
\setlength{\tabcolsep}{3pt}
\renewcommand{\arraystretch}{1.1}
\begin{table}[!t]
\centering
\resizebox{\linewidth}{!}{%
\centering
% \begin{table}[!th]
% \centering
% \color{black}
\begin{tabular}{lcccc}
\hline
& & \multicolumn{2}{c}{Video-Level} & \multicolumn{1}{c}{Frame-Level}\\\cline{3-4}
Prompt & Setting & Video-LLaVA & AffectGPT & Video-LLaVA\\ \hline
Simple             & ZS   & 0.2827 & 0.2933 & 0.4416\\ 
Definition 1 Only  & ZS   & 0.3326 & 0.3772 & 0.4456\\ 
Definition 2 Only  & ZS   & 0.3772 &  0.3772 & 0.1651\\ 
Transcript + Simple & ZS & NA &  0.6016 & NA\\ 
Transcript + Def 1 & ZS   & 0.6341 & 0.6436 & \textbf{0.4535}\\ 
Transcript + Def 1 & FT   & \textbf{0.7142} & \textbf{0.7229} & NA\\ 
Transcript + Def 2 & ZS   & 0.3945 & 0.6462 & 0.1849\\ 
\hline  
\end{tabular}
}
\caption{\avgfonescore performance of video and frame level predictions using MLLMs. 
}
% \label{tab:mllm-video}
\label{tab:mllm-video-frame}
\vspace{-15pt}
\end{table}
}

The performance of MLLMs with zero-shot inference is substantially influenced by the query prompt. As observed from Table~\ref{tab:mllm-video-frame}, simply asking the model to predict emotion based on the visual modality performs the worst, whereas adding only the definition of A/H in the query prompt improves the model's performance. The best results in all cases are obtained with the introduction of the text transcript of the video in the query prompt. We conjecture that this happens for two reasons: i) the textual modality serves a significant role in the identification of the A/H, and ii) the current MLLMs' performance is heavily reliant on the textual modality. This aligns with the overall structure of traditional MLLMs that are built upon well-trained LLMs with the addition of a visual encoder like ViT, which is used to encode the visual information that is fed to the LLM for downstream tasks. Intuitively, the performance should increase with careful fine-tuning on the \bah dataset. 
Further, the idea of textualizing the audio and visual modalities explored in~\cite{richet-abaw-24} can be suitable for our task, where the audio and visual modalities essentially summarize the cues detected in the corresponding modalities. Particularly for tasks like subtle emotion recognition or the detection of A/H, where cross-modal inconsistency has to be considered, textualizing the audio and visual modalities can be done to adequately exploit the reasoning abilities of LLMs.

% =======================================================================
%                  RECOMMENDATIONS
% =======================================================================
\section{Recommendations for Future Work} \label{sec:recommendations}

The newly introduced A/H recognition task requires the model to detect conflicting affects across and within modalities. Standard multimodal models are trained to yield predictions aligned with the output supervision. This automatic training may focus on learning label patterns and miss acquiring a mechanism to understand affect conflict. To build more interpretable A/H recognition systems, we recommend a 2-level framework. The first level should focus on modeling affect per modality in an independent way. Off-the-shelf pretrained sentiment analysis models could be used~\cite{sharma25}. This first level should be separated from A/H since we cannot detect it at the modality level yet, at least for the cross-modality case. At the second level, a dedicated fusion mechanism should be used to assess whether there is conflicting affect across modalities to make a decision. This module does deeper work than simply feature fusion, as is commonly done. It should acquire an understanding of affect conflict to be able to detect it. Such a modular and interpretable framework allows introducing priors about affect conflicts. 

Statistics extracted from annotator cues in \bah could be leveraged to constrain the model to find conflicts across modalities. A specialized temporal modeling approach could be used to detect within-modality conflict based on the output of the first level. Statistics from the A/H segment durations could be considered, as A/H happens briefly.  Context is also important to detect within-modality cases. A segmented body could help as well since it contains important cues and less noise compared to a full frame. Our results showed that standard multimodal models and fusion techniques yield modest performance. Future work should focus on designing specialized frameworks for A/H recognition.

%####################################
\section{Conclusion}
\label{sec:conclusion}
Automating digital health interventions, especially through ML, has recently gained significant attention.
A/H plays a primary role in delaying, avoiding, or abandoning health behaviour changes. Automatic and accurate A/H recognition is therefore critical for the personalization and cost-effectiveness of digital behaviour change interventions.
We explored, in this work, the application of deep learning models for the task of A/H recognition in videos. We cover three learning scenarios: supervised learning, unsupervised domain adaptation for personalization, and zero-shot inference via LLMs. Our experiments are conducted on the unique and recently published dataset \bah for A/H recognition. Our results showed that existing models have limited performance. This suggests that more adapted multi-modal models are required for accurate A/H recognition. Moreover, better ways to model spatio-temporal and modality fusion are necessary to detect within- and cross-modality conflicts.

\subsubsection*{Acknowledgements}
This work was supported in part by the Fonds de recherche du Québec – Santé, Natural Sciences and Engineering Research Council of Canada, Canada Foundation for Innovation, and Digital Research Alliance of Canada.

% \newpage
\section*{Ethical Impact Statement}

\noindent \textbf{Research with human subjects.} This work does not include a new dataset. In all our experiments, we used the publicly available \bah~\cite{gonzalez-26-bah} dataset. The dataset was collected in an ethical manner after acquiring approval from two ethics committees from two universities. All participants provided their formal consent to be part of the dataset and the research. The dataset does not contain private information that can identify participants. The dataset is video-based. It contains recordings of participants answering questions with their visual and audio data recorded in the participant environment. The dataset does not contain offensive content.

\noindent \textbf{Potential negative impact.} This work is intended to develop methods for A/H recognition to help participants modify unhealthy behaviours and therefore improve their well-being. However, there is a potential risk of using this application for what is deemed bad and harmful in society. Detecting A/H can be used to rethink interventions and adapt tools that might influence people's behaviours and attitudes. 

\noindent \textbf{Limits of generalizability.} Our experiments are limited to health interventions based on question-answer, reflecting the \bah dataset. We do not claim that our results safely extend to all A/H scenarios. While this dataset covers a diverse population (e.g., age, sex, ethnicity), with very limited societal bias, it does not cover all cultures. Such an aspect can affect how a person manifests A/H.

\FloatBarrier
% \clearpage

\bibliographystyle{abbrv}
{
\bibliography{main}
}

\end{document}